\documentclass[sigconf]{acmart}
\AtBeginDocument{%
  \providecommand\BibTeX{{%
    \normalfont B\kern-0.5em{\scshape i\kern-0.25em b}\kern-0.8em\TeX}}}

\setcopyright{none}
\copyrightyear{}
\acmDOI{}
\acmISBN{}
\acmConference[Data Science Seminar]{Data Science Seminar}{2019}{Uni Passau}

\begin{document}
\title{Comparison of Neuronal Attention Models }

\author{Mohamed Karim BELAID}
\email{belad01@ads.uni-passau.de}
\affiliation{%
  \institution{University of Passau}
  \city{Passau}
  \state{Germany}
  \postcode{94032}
}

\begin{abstract}

  Recent models for image processing are using the Convolutional neural network (CNN) which requires a pixel per pixel analysis of the input image.
  This method works well. However, it is time-consuming if we have large images.
  To increase the performance, by improving the training time or the accuracy, we need a size-independent method.
  As a solution, we can add a Neuronal Attention model (NAM).
  The power of this new approach is that it can efficiently choose several small regions from the initial image to focus on.\newline
  The purpose of this paper is to explain and also test each of the NAM's parameters.
\end{abstract}

\maketitle

\section{Introduction}
Starting with Yarbus in 1950, Human perception has been the focus of many research works \cite{Yarbus67,Hiroyuki01,Mary05} . One of the goals was to find the main components which represent the human vision.
One of the recent theory models the entire process with these three rules 
\cite{Mary05}: First, the eye analyses small regions of the environment in a sequential way. Second rule, the next target depends on the content of the current one. Third rule, the results of this active search are transformed into a reward and also the next step is chosen so that the potential reward is the maximum.
The Recurrent Neural Network introduced in this paper \cite{DBLP:journals/corr/MnihHGK14} mimics the eye movement as described above.  

Image captioning is one of the applications of the Visual Attention Model.
\begin{figure}[h]
  \centering
  \includegraphics[width=\linewidth]{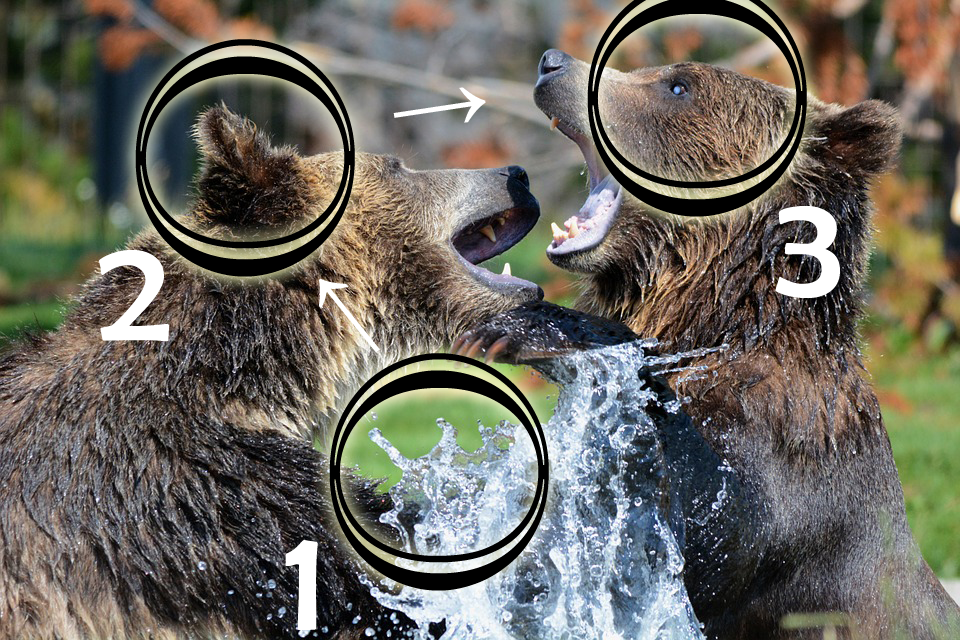}
  \caption{3 Glimpses are used to find a suitable title to this image, ``2 Grizzly bears playing in water''.}
  \Description{2 Grizzly bears playing in water}
\end{figure}
The purpose of this paper is to go through the parameters introduced by this method and analyze its impact on the accuracy and the training duration.

\section{Related Work}
During the last 20 years, different approaches to Model Attention have been tested \cite{Itti12}. Attention can be expressed in different ways: Either the model detects and focuses on pertinent data points to learn (Item-wise) or it focuses on a segment of the example (Location-wise)

This paper is based on the Recurrent Attention Model (RAM) introduced by Mnih et al. (2014)\cite{DBLP:journals/corr/MnihHGK14}. The studied model is a location-wise Attention model. the latter can be improved by adding a spatial transformation network (STN)\cite{2015arXiv150602025J}. STN enable the model to generalize over transformed data (e.g. scaled, translated and rotated images)
It has been shown that RAM models outperform CNN models. As an example, the whole sequence recognition error rate on
enlarged multi-digit SVHN is 50\% for a 10 layer CNN model. on the other hand, the test error rate of the forward-backward D-RAM fine-tuning model is 4.46\% \cite{Itti12}.\newline
This paper resumes the work done by Tianyu\cite{Tianyu17}. A clear implementation can be found here \cite{Mohamed19}. The code is tested with more than 30 different configurations. The detailed results are available in this repository \cite{Mohamed19} and can be visualized with Tensorboard.

\section{Dataset}

The MNIST dataset challenge \cite{lecun-mnisthandwrittendigit-2010} has been solved with different approaches with an accuracy approaching the 100\% \cite{BAYDILLI18, Keras18}.
This dataset contains 70'000 grayscale images of fixed-size (100 x 100 pixels) in which each image represents only one digit.
To make it more challenging, an augmented MNIST was created. An additive noise was introduced and the digits were rotated then translated.

\section{Neuronal Attention model}
\subsection{Analyzed Model}

\begin{figure*}
  \centering
  \includegraphics[width=\textwidth]{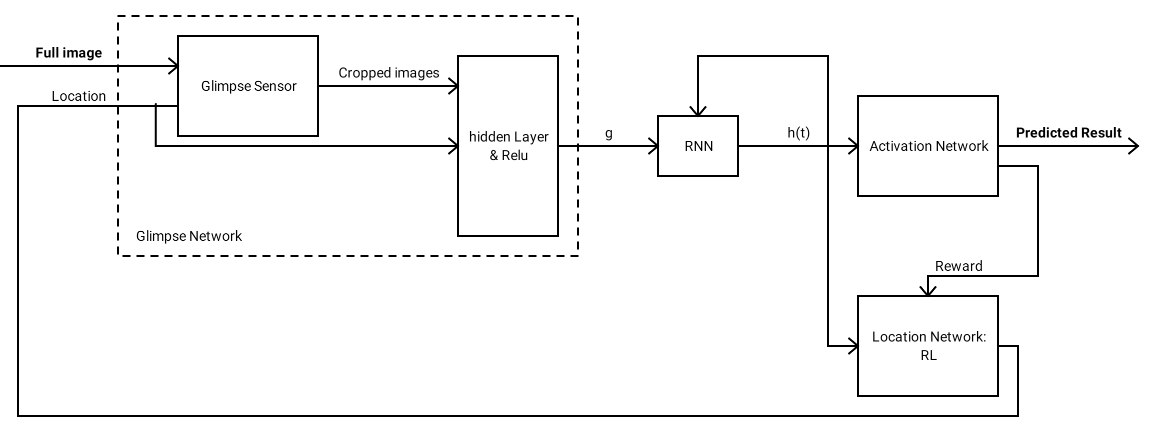}
  \caption{Neural Attention Model's flow chart.}
  
  \label{fig:flowchart}
  \Description{NAM is composed of 5 main elements.}
\end{figure*}

The model studied was presented by Volodymyr et al. in 2014\cite{DBLP:journals/corr/MnihHGK14}. It has 5 components:
\begin{itemize}
\item The \textbf{Glimpse sensor} is the kernel of the final solution. It takes two inputs: the full image and a location. The output is a retina-like image. The image has a precise center with blurry edges.
\item The \textbf{Glimpse network} combines this output with the location through a hidden layer to create the vector $g$.
\item The vector $g$ goes through a \textbf{Recurrent network}. This network filters the most important arguments and gives as output the vector $h_t$. 
\item The \textbf{Activation network} uses $h_t$ to predict the digit.
\item In parallel, the \textbf{Location network} uses $h_t$ and a reward provided by the activation network to predict the next location.
\end{itemize}
The flowchart in figure \ref{fig:flowchart} summarizes the studied model. More details about each component can be found here \cite{blog17}

\subsection{Environment}

The training is performed using the Keras framework with TensorFlow backend \cite{199317}  .
To obtain a comparable result, we will use only one GPU-equipped environment ( GeForce GTX1060 ) with a computation capability of 6.1 . The detailed performance of the machine was appended as meta-data to each log for the Tensor-board visualization.

\subsection{Baseline}
Table~\ref{tab:baselineConfig}
represents the baseline for all training scenarios.
\begin{table}[h]
  \caption{Baseline configuration for all training.}
  \label{tab:baselineConfig}
  \begin{tabular}{lc}
    \toprule
    Parameters&Value\\
    \midrule
    Number of glimpse & 4\\
    Number of scales&4\\
    Sensor bandwidth& 12x12 pixels\\
    Noise standard deviation &0.22\\
    Batch size&128\\
    Number of epoch&100\\
    Optimizer&Adam\\
    Learning rate&$10^{-3}$\\
    decay &0.97\\
  \bottomrule
\end{tabular}
\end{table}
\newline In the following figures, the baseline scenario is the one colored in orange (Figure \ref{fig:glimpsetime} \ref{tab:scalestime} ...).

\section{Hyper-parameters description}
\subsection{Methodology}
In this section, the most important hyper-parameters will be described one by one. Then, a linear search on this feature will be performed in order to reveal its impact on the accuracy and the training time.\newline
To follow the evolution of the accuracy a test is made after each mini-batch execution. For this reason, the total training time is much longer. Therefore, the obtained numbers are only useful for relative comparison.
And due to this long training time, some scenarios have been executed with fewer epochs (25, 50 or 75). This change is always mentioned in the figures.
\subsection{Model-Based Hyper-parameters}
\subsubsection{Number of glimpses}

To understand the content of the image, the Neural Attention Model has to analyze different regions. This hyper-parameter describes the number of ``shots'' analyzed from one image. 
In this example (Figure 1) we can detect the 2 animals and the water with a minimum of 3 glimpses. Of course, some other glimpses will be used to find these accurate elements in the image.
Street View House Number (SVHN) transcription is one of the applications that shows the importance of this parameter.

\begin{figure}[h]
  \centering
  \label{fig:svhn}
  \includegraphics[width=\linewidth]{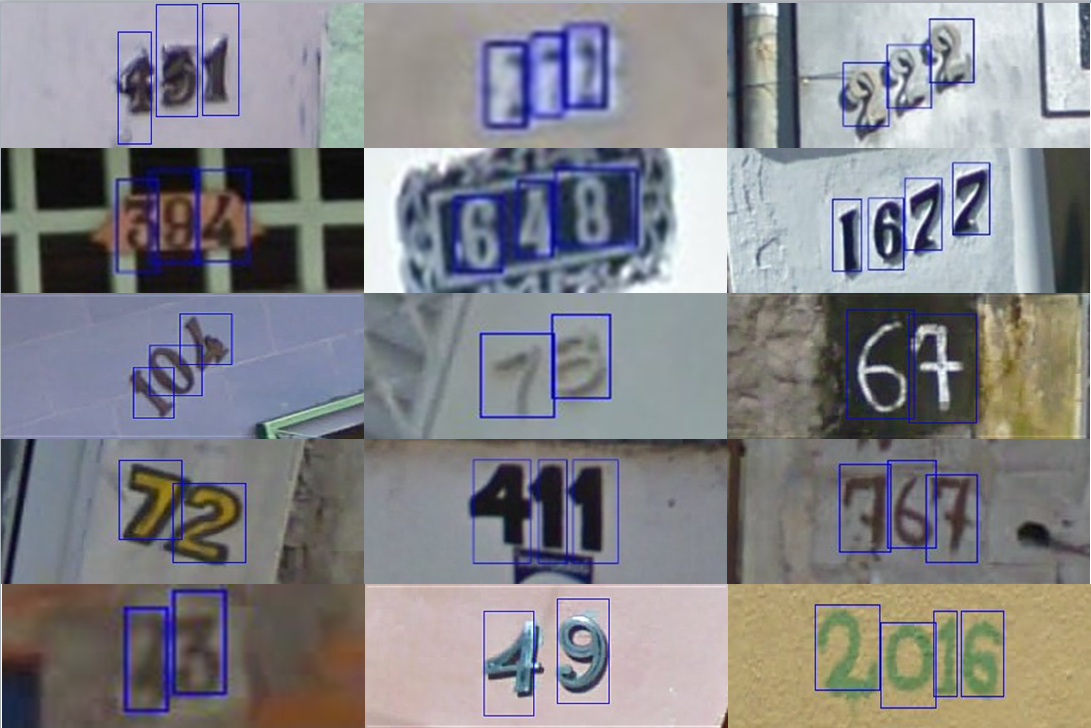}
  \caption{Examples of cropped images from the SVHN database. Each digit is rotated and has a different relative position. Some images are also blurred\cite{svhn11}.}
  \Description{examples of house numbers from the SVHN Database}
\end{figure}

To summarize the SVHN problem, we need to retrieve the number of houses from a high-resolution image.
A CNN with 11 hidden layers can recognize the sequence of numbers with high efficiency. After 6 days the model reaches an accuracy of around 95\% \cite{42241}
Different approaches based on Deep Recurrent Attention Model solved the same problem with slightly the same accuracy ($\pm2\%$) but with less computation effort. The training is 4 to 40 times faster, depending on the configuration \cite{blog17}.

\begin{figure}[h]
  \centering
  \includegraphics[width=0.8\linewidth]{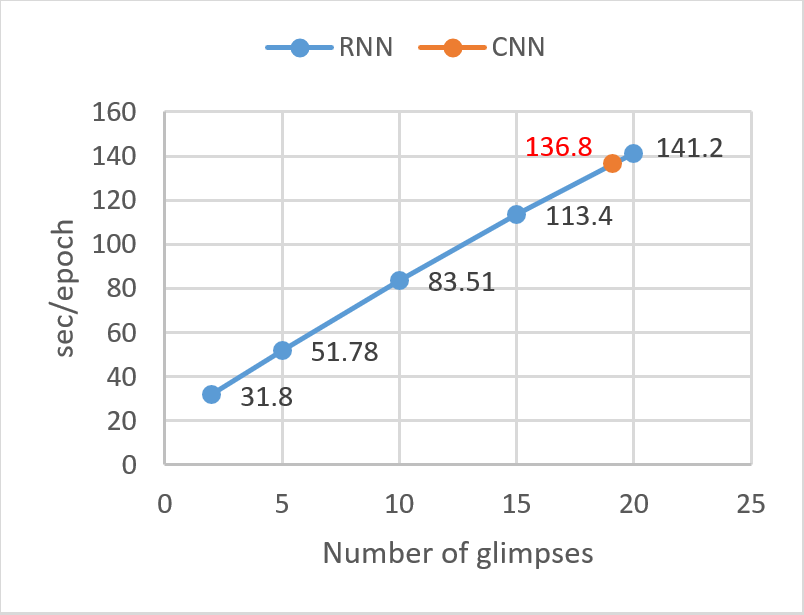}
  \caption{The training time on MNIST database using the baseline scenario and varying only the number of glimpses.}
  \Description{the number of glimpses has a linear time cost}
  \label{fig:NumberOfGlimpses}
\end{figure}

Figure \ref{fig:NumberOfGlimpses} shows that the execution time is linear to the number of glimpses.

To obtain a fast training rate, we need to limit the number of glimpses to 15. Otherwise, a CNN model will be more efficient. \cite{Keras18}
Generally, 4 to 8 glimpses are enough to obtain good results for the MNIST dataset.

\begin{figure}[h]
  \centering
  \label{fig:StatNbOfGlimpses}
  \includegraphics[width=\linewidth]{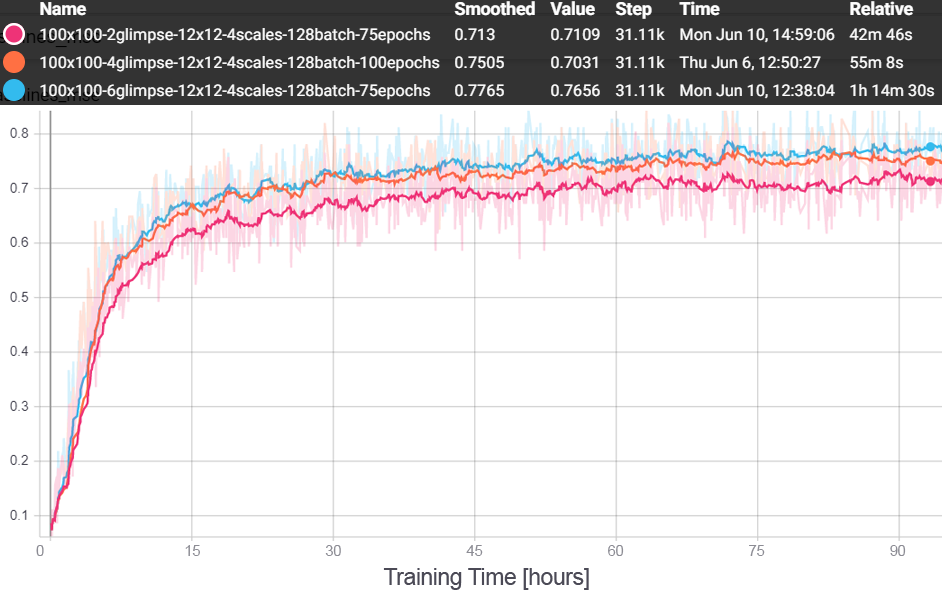}
  \caption{Evolution of the average accuracy for 100 epochs. 3 different numbers of glimpses were tested.}
\end{figure}
The number of glimpses impacts, in a linear way, not only the training but also the testing time.
It is therefore important to choose it carefully if we have a real-time application.

\begin{figure}[h]
  \centering
  \includegraphics[width=\linewidth]{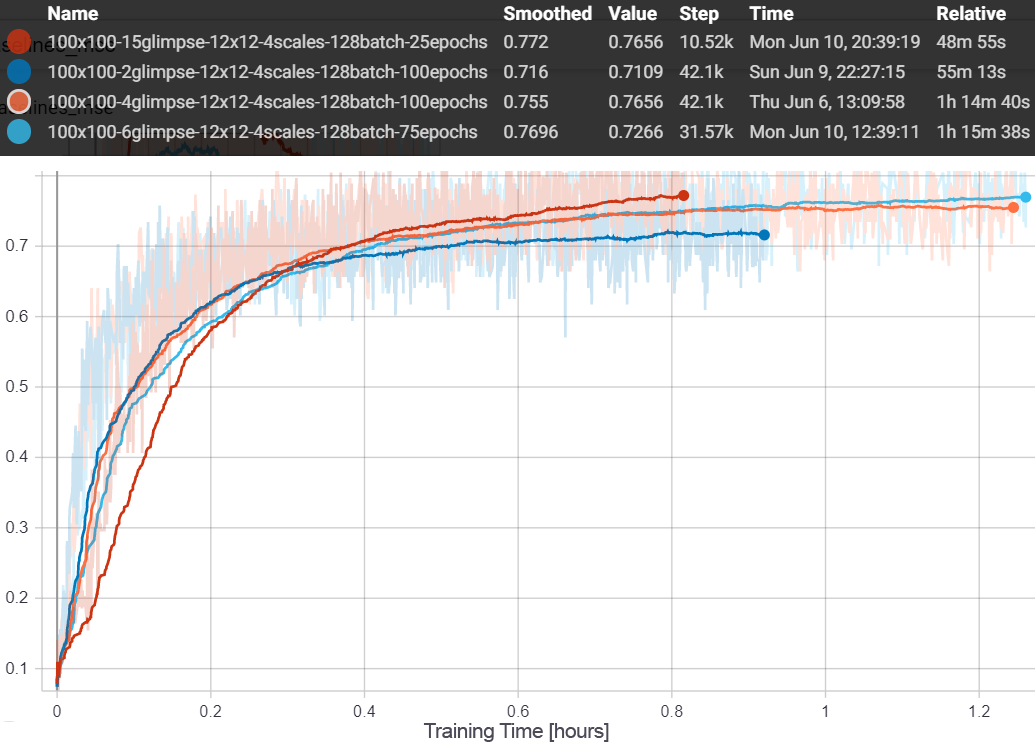}
  \caption{Evolution of the average accuracy through  time while varying the number of glimpses.}
  \label{fig:glimpsetime}
\end{figure}

In figure \ref{fig:glimpsetime}, the X-Axis represents the training time. In the first 30 minutes, the accuracy is inversely proportional to the number of glimpses. After a longer training time, it becomes proportional.\newline
So, the number of glimpses can not be fixed to a high value for all applications: As for devices with limited computation capability, or for real-time applications, we should use a small value (2 to 6 for MNIST). Thus, we obtain a lighter model in a small amount of time.
But if we are looking for the best accuracy we will have to use a larger value (8 to 16 for MNIST).

\subsubsection{Sensor size}
The $Sensor\ size$
represents the number of weights
in the Glimpse Network.
The analysis of the $Sensor\ size$ bring us to the analysis of the $number\ of\ scales$ and the $bandwidth$ size since,\newline
$Sensor\ size\ =\ bandwidth^2 \cdot number\ of\ scales$

\subsubsection{Scales}
To focus the attention on a certain region of the image, we crop around a certain location.
\begin{figure}[h]
  \centering
  \includegraphics[width=\linewidth]{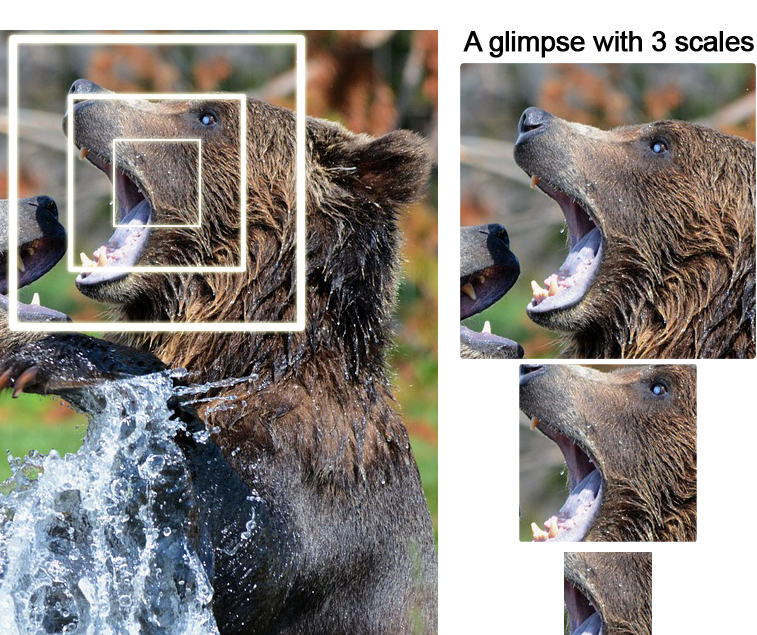}
  \caption{Example of output of the Glimpse sensor for scales=3.}
  \label{fig:bearscales}
\end{figure}

The Glimpse Sensor can output a different number of scales. 4 is the baseline value used to train MNIST. An accuracy of 80\% is obtained in less than 3 hours.

\begin{table}
  \caption{Evolution of the training time while varying the number of scales.}
  \label{tab:scalestime}
  \begin{tabular}{cc}
    \toprule
    Number of scales&Second / epoch\\
    \midrule
    2&36.6\\
    4&46.1\\
    6&115.5 \\
    8&368 \\
    
  \bottomrule
\end{tabular}
\end{table}

\subsubsection{sensor bandwidth}
The sensor bandwidth parameter represents the resolution of all scales.

\begin{figure}[h]
  \centering
  \includegraphics[width=\linewidth]{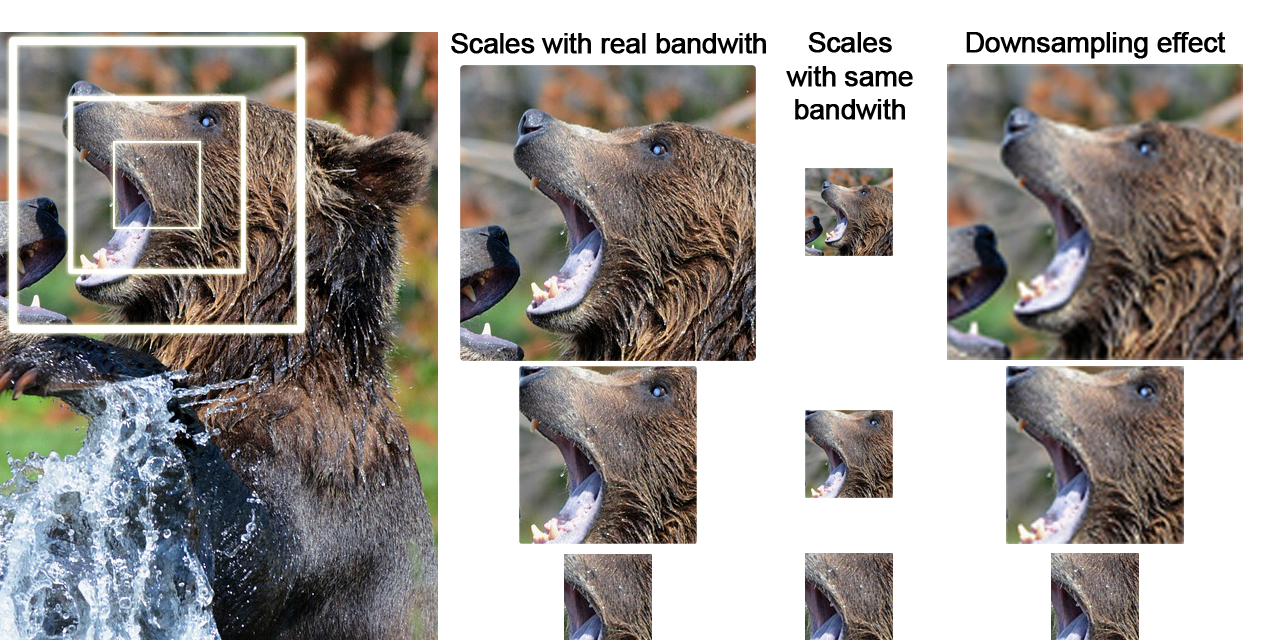}
  \caption{Example of 3 scales with a bandwidth of 88x88 $px$}
  \label{fig:downsample}
\end{figure}
In Figure \ref{fig:downsample} the scales are downsampled to 88x88 pixels. Thus, the blurring effect of the eye's vision is reproduced.\newline
MNIST dataset is composed of 100 x 100 grayscale images. The baseline value used for downsampling is 12x12 pixels.

\begin{figure}[h]
  \centering
  \includegraphics[width=\linewidth]{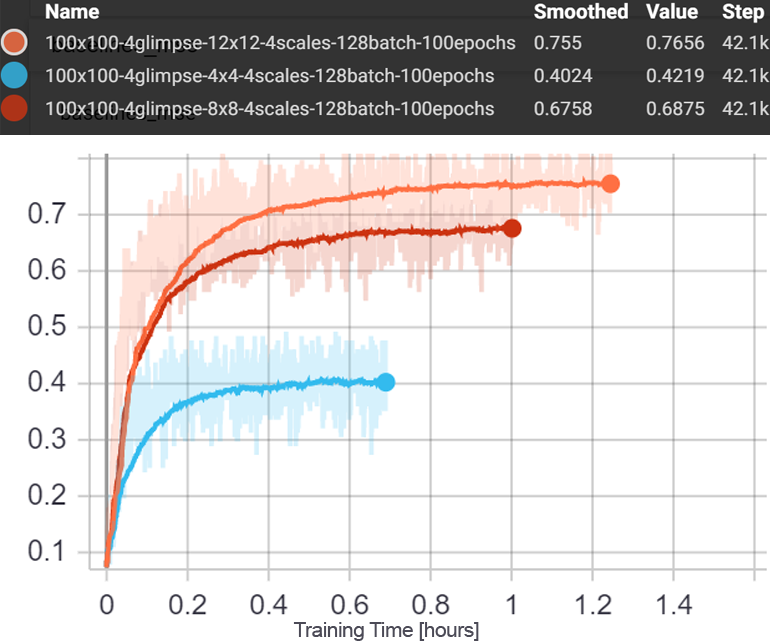}
  \caption{Evolution of the accuracy through time while varying the bandwidth.}
  \label{fig:bandwidthtime}
\end{figure}

Unlike the number of glimpses, the sensor bandwidth is always proportional to the average accuracy. There is no need to select different values for low computation devices. This can be observed in Figure \ref{fig:bandwidthtime}.

\begin{figure}[h]
  \centering
  \includegraphics[width=0.85\linewidth]{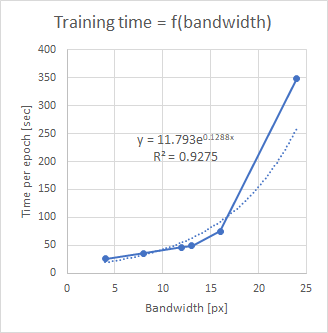}
  \caption{Evolution of the training time while varying the bandwidth}
  \label{fig:bandwidthtimeexcel}
\end{figure}

Let's consider Figure \ref{fig:bandwidthtimeexcel}. For high values, we quickly encounter the problem of curse of dimensionality since that for a bandwidth of 12 pixels, for example, we create $( 12^2 \cdot Number\ of\ scales )$ weights.
\subsubsection{Scales vs sensor bandwidth}

\begin{figure}[h]
  \centering
  \includegraphics[width=\linewidth]{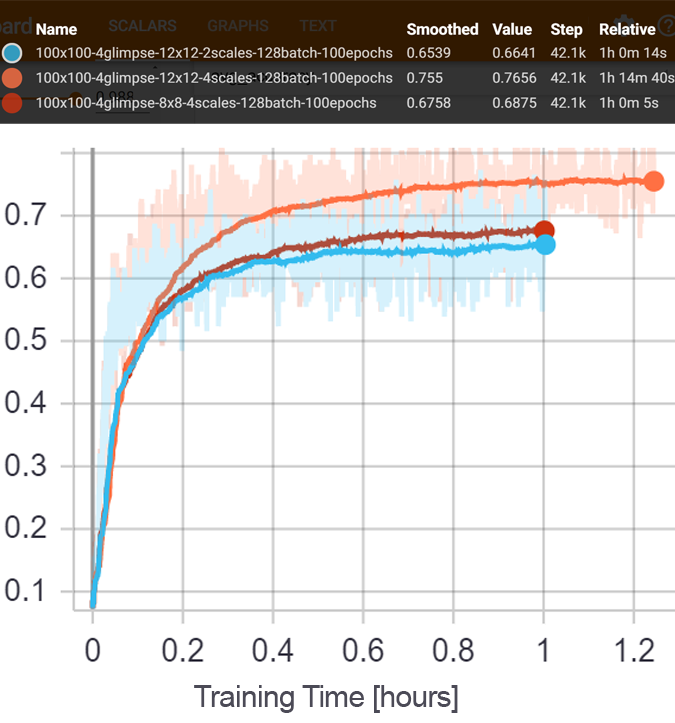}
  \caption{Comparing the effect of bandwidth and number of scales.}
  \label{fig:bandwidthVSscales}
\end{figure}
Balancing between scales and sensor bandwidth will lead to the same computation cost, but not the same performance. The number of scales is more important than the number of pixels per scale. This is shown by comparing the red and blue scenarios in Figure \ref{fig:bandwidthVSscales}. The orange scenario is the baseline.

\subsubsection{Noise Standard Deviation}
The location network follows Reinforcement Learning rules: First, a reward matrix is received from the Activation Network. A reward equal to 1 is assigned to a pixel if the Activation Network might use it to predict a number. Otherwise, the reward is null. Secondly, a gradient is calculated from this 2-dimensional matrix, the glimpse Sensor will move in the direction of the gradient. The goal is to maximize the reward by maximizing the number of useful pixels observable in the glimpse. Thirdly, Gaussian noise is added to the predicted next location. Thanks to this optimization, we can escape local minima and explore more useful locations\cite{ruder16}.
In this part, we study the effect of the noise's standard deviation on the reward.

\begin{table}
  \caption{Results for different standard deviation configurations.}
  \label{tab:stdResult}
  \begin{tabular}{lcc}
    \toprule
    Std Value&Accuracy&Time\\
    \midrule
    0.10&0.65&1h16\\
    0.22&0.75&1h16\\
    0.4&0.83&1h48\\
    0.6&0.84&2h15\\
    1.2&0.79&2h40\\
  \bottomrule
\end{tabular}
\end{table}
First, we notice that the training time is affected by the standard deviation value. The training time can even double. Second, the accuracy is at its maximum around a certain range ( 0.4 to 0.6 ). For a small std value, the effect of random noise is almost nonexistent. It does not allow exploration.
On the other hand, high values cancel the deterministic movement suggested by gradient descent: by increasing the standard deviation, predicting the next location will tent to become a simple random search.
To summarize, there is a trade between the movement suggested by gradient descent and the additive random movement. The objective is to let the gradient descent guide the glimpse's movement, then, in a second step, use Gaussian noise to avoid local minima and explore.

\subsection{Train-Based Hyper-parameters}
\subsubsection{Batch size}

\begin{figure}[h]
  \centering
  \includegraphics[width=\linewidth]{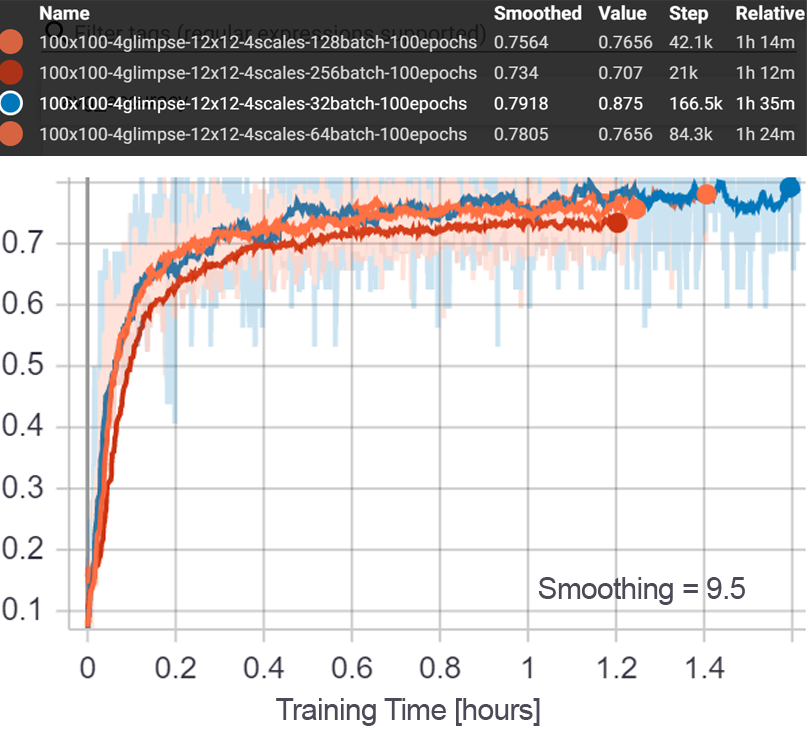}
  \caption{Effect of the batch size on the accuracy and the training time.}
  \label{fig:batchsize}
\end{figure}

As we can see in Figure \ref{fig:batchsize}, increasing the batch size from 32 to 256, decreases the training time by 25\% and decreases the accuracy by 7\%.
We notice also that low batch size involves a high variance of the accuracy.
\subsubsection{Epoch}
Decreasing the batch size is like applying early stopping: It does not change the shape of the figure.
For a batch size of 128, 50 epochs are enough to reach convergence, independently of the rest of the configuration. The baseline was 100 epochs to ensure that we reached the convergence for each configuration.\newline
Decreasing the batch size will delay the convergence. For this reason, a batch size of 32 would require more epochs.

\subsubsection{Optimizer}
The Adaptive Moment estimation (Adam) is an optimization algorithm. It extends the stochastic gradient descent \cite{Kingma2014AdamAM} by calculating different learning rates, each adapted to its parameter. Adam Optimizer combines both AdaGrad and RMSProp. More about these methods can be found here\cite{Jason2017}.

For CNN models, the Adam Optimizer showed the best accuracy and loss \cite{BAYDILLI18} on the MNIST database. For this reason, it was used in the baseline configuration.\newline
Changing the Optimizer is not an easy task: It requires picking the right learning rate. As each optimizer learns only in a certain range \cite{bestLearningRate}.
For example, using the AdaGrad Optimiser with a starting learning rate of 0.22 will result in an accuracy of 0.1 for more than 30 epoch. The model is therefore equivalent to the baseline random search.\newline
Based on a pre-study\cite{bestLearningRate} of the optimal ranges, different optimizers were tested. 

\begin{table}
  \caption{Results for different optimizers.}
  \label{tab:optimizerResult}
  \begin{tabular}{lccc}
    \toprule
    Optimizer&Learning rate&Accuracy&Time\\
    \midrule
    AdaGrad&0.2&0.75&1h22 \\
    Adam&$10^{-3}$&0.75&1h16 \\
    AdaDelta&5.0&0.78&1h20 \\
    RMSProp&0.0009&0.79&1h25 \\
  \bottomrule
\end{tabular}
\end{table}

Based on these results, We notice that the choice of the Optimizer does not impact the training time significantly ($\pm$ 5 min, equivalent to $\pm$7\%).\newline

\begin{figure*}[!t]
  \centering
  \includegraphics[width=\textwidth]{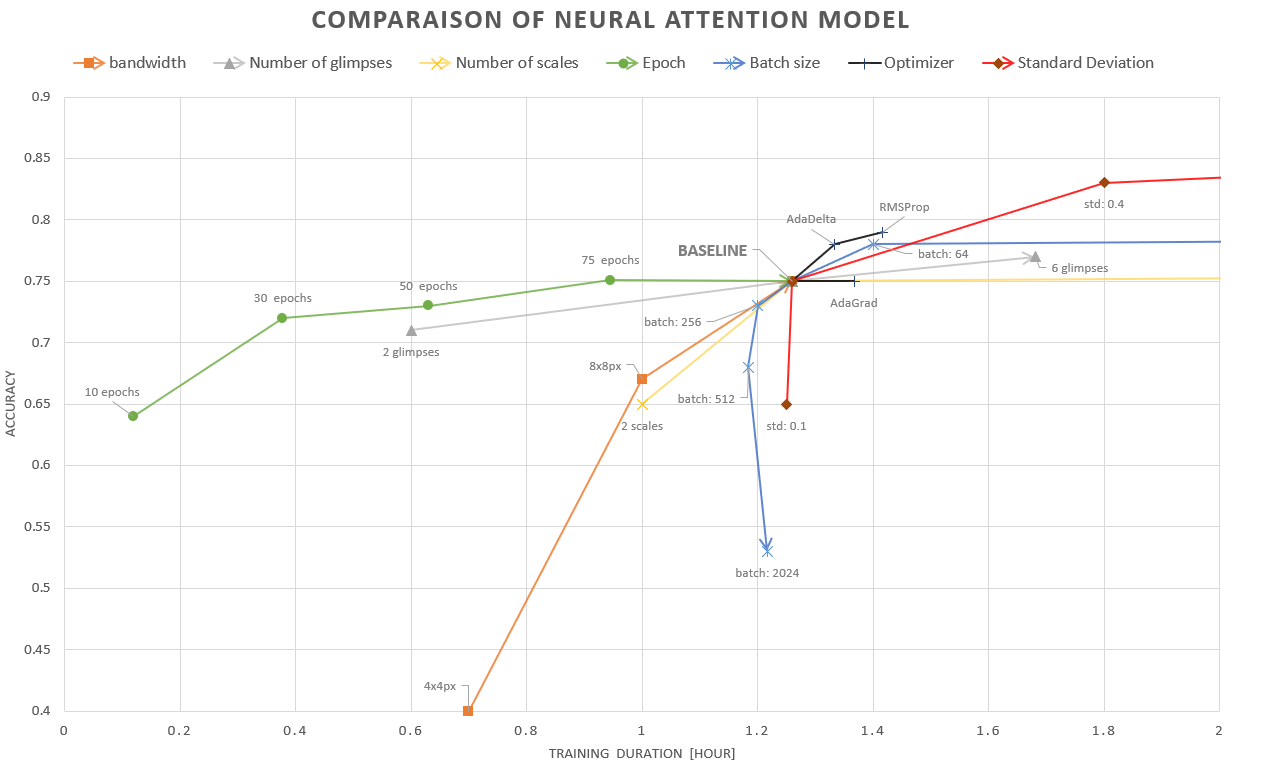}
  \caption{Different configurations of Neural Attention Model.}
  
  \label{fig:conclusion}
  \Description{23 configurations of Neural Attention Model.}
\end{figure*}
Concerning the accuracy, all optimizers result in slightly the same accuracy (between 0.75 and 0.8) There is no reason to pick Adam. But for hyper-parameter optimization, a deeper analysis of the learning rate needs to be performed to distinguish the best configuration.

\subsection{Other parameters}
For certain reasons, some parameters were not tested in this work:
\begin{itemize}
\item Previous research line considered only one channel of the image since the images of MNIST are grayscale. It would be possible to test this parameter with the SVHN database.
\item The Location Net. incorporates a Reinforcement Learning model. It includes different parameters --- like the Monte Carlo sampling --- which are not tested in this paper.
\item Introducing a spatial transformer network would turn this solution to what is called a ``soft'' Attention Model. This layer was explained deeply in this article \cite{blog17}.
\end{itemize}

\section{Conclusion}

We described a novel digit recognition technique, based on an attention mechanism. Considerable computation effort was saved by focusing the analysis on small regions of the image. This mechanism brings more than 10 hyper-parameters that directly affect the learning and prediction processes.
The MNIST database specification enabled us to test 7 of them. The graphic n$^\circ$\ref{fig:conclusion} is a  summary figure describing the main findings of the current paper. \newline
As a conclusion, all parameters have tight intervals where they provide good accuracy. But with wrong values, the learning process would be unnecessarily prolonged.\newline
The sensor bandwidth parameter has the highest effect on accuracy. Increasing it, will add 30 minutes to the training time and the accuracy will jump by 35\%.
Other parameters affect less the accuracy ($\pm$10\%) but double the training time.

\begin{acks}
This paper and the research behind it would not have been possible without the exceptional support of my supervisors, J\"org Schl\"otterer and Prof. Dr. Michael Granitzer. I would like to show my gratitude to Dorra Elmekki, for her comments on an earlier version of the paper,  although any errors are our own and should not tarnish the reputations of these esteemed persons.
\end{acks}

\bibliographystyle{ACM-Reference-Format}
\bibliography{sample-base}
\end{document}